\documentclass[10pt,letterpaper]{article}
\usepackage[margin=0.88in,top=0.78in,bottom=0.82in]{geometry}
\usepackage{times}
\usepackage[round,authoryear]{natbib}
\usepackage{fancyhdr}
\usepackage{titlesec}

\usepackage[T1]{fontenc}
\usepackage[utf8]{inputenc}
\usepackage{microtype}
\usepackage{inconsolata}
\usepackage{graphicx}
\usepackage{booktabs}
\usepackage{hyperref}
\usepackage{url}
\usepackage{amsfonts}
\usepackage{dsfont}
\usepackage{nicefrac}
\usepackage{xcolor}
\usepackage{amsmath}
\usepackage{amssymb}
\usepackage{mathtools}
\usepackage{amsthm}
\usepackage{xspace}
\usepackage{xurl}
\usepackage{makecell}
\usepackage{enumitem}
\usepackage[capitalize,noabbrev]{cleveref}
\usepackage{subcaption}
\usepackage{algorithm}
\usepackage{algpseudocode}
\usepackage{multirow}
\usepackage{tabularx}
\usepackage{array}
\usepackage{wrapfig}
\usepackage{colortbl}
\usepackage{float}
\usepackage{listings}
\usepackage[normalem]{ulem}
\usepackage[most]{tcolorbox}
\usepackage{pifont}
\usepackage{siunitx}
\usepackage{tikz}
\usetikzlibrary{arrows.meta,positioning,fit}
\hypersetup{hidelinks}

\definecolor{EKTAccent}{HTML}{397C82}
\colorlet{EKTRow}{EKTAccent!6}

\definecolor{groundblue}{HTML}{0072B2}
\definecolor{spatialteal}{HTML}{009E73}
\definecolor{actionorange}{HTML}{D55E00}
\definecolor{stressamber}{HTML}{E69F00}
\definecolor{lossred}{HTML}{A33A3A}
\definecolor{softgray}{HTML}{F2F4F5}
\definecolor{midgray}{HTML}{6B7075}
\definecolor{groundwash}{HTML}{E8F3F9}
\definecolor{spatialwash}{HTML}{E8F5F0}
\definecolor{actionwash}{HTML}{FAEEE9}
\definecolor{losswash}{HTML}{F8ECEC}

\hypersetup{hidelinks}

\def\method{\emph{EKT-Router}\xspace}

\newcommand{\emptyk}{\varnothing_{\mathcal K}}

\definecolor{RouterTeal}{HTML}{397C82}
\definecolor{RouterBlue}{HTML}{526B92}
\definecolor{RouterInk}{HTML}{1F3049}
\titleformat{\section}{\large\bfseries\color{RouterInk}}{\thesection}{0.7em}{}
\titleformat{\subsection}{\normalsize\bfseries\color{RouterInk}}{\thesubsection}{0.7em}{}
\titlespacing*{\section}{0pt}{2.2ex plus .5ex}{1.0ex}
\titlespacing*{\subsection}{0pt}{1.6ex plus .4ex}{0.7ex}
\hypersetup{colorlinks=true,linkcolor=RouterTeal,citecolor=RouterTeal,urlcolor=RouterBlue,
 pdftitle={EKT-Router: Expert–Evidence Alignment for Embodied Vision–Language Models},
 pdfauthor={Weilin Ruan, Yingjie Xu, Yangkai Wei, Bohan Zhou, Shuaibo Li, Harold Haodong Chen, Kaiwen Zhou, Yinchuan Li, Ying-Cong Chen, James Cheng}}
\renewenvironment{abstract}{\begin{tcolorbox}[colback=RouterTeal!4,colframe=RouterTeal!12,boxrule=.4pt,arc=3pt,left=10pt,right=10pt,top=8pt,bottom=8pt]\small\noindent\textbf{\textcolor{RouterTeal}{ABSTRACT}}\quad}{\end{tcolorbox}}
\begin{document}
\thispagestyle{empty}
\begin{center}
{\resizebox{\linewidth}{!}{\bfseries\color{RouterInk}\textcolor{RouterTeal}{EKT-Router:} Expert--Evidence Alignment for Embodied Vision--Language Models}}\par\vspace{12pt}
{\normalsize\bfseries
Weilin Ruan\textsuperscript{1,2}, Yingjie Xu\textsuperscript{2,3}, Yangkai Wei\textsuperscript{2}, Bohan Zhou\textsuperscript{1,2},\\[3pt]
Shuaibo Li\textsuperscript{1}, Harold Haodong Chen\textsuperscript{3,4}, Kaiwen Zhou\textsuperscript{2},\\[3pt]
Yinchuan Li\textsuperscript{2}, Ying-Cong Chen\textsuperscript{3}, James Cheng\textsuperscript{1,*}}\par\vspace{7pt}
{\small\textsuperscript{1}The Chinese University of Hong Kong\quad
\textsuperscript{2}Knowin.AI\\[2pt]
\textsuperscript{3}The Hong Kong University of Science and Technology (Guangzhou)\\[2pt]
\textsuperscript{4}The Hong Kong University of Science and Technology}\par\vspace{9pt}
{\small\sffamily\bfseries
\href{https://knowin-29.github.io/KnowinRouter.github.io/}{Project}\quad\textcolor{midgray}{/}\quad
\href{https://github.com/Knowin-29/EKT-Router}{Code}\quad\textcolor{midgray}{/}\quad
\href{https://huggingface.co/datasets/Knowin-29/EKT-Router}{Data}\quad\textcolor{midgray}{/}\quad
\href{https://huggingface.co/Knowin-29/EKT-Router}{Model}}\par\vspace{5pt}
{\footnotesize\sffamily\textcolor{RouterTeal}{CONTACT}\quad\href{mailto:rwlinno@gmail.com}{rwlinno@gmail.com}\quad\textcolor{midgray}{* corresponding author}}
\end{center}
\begin{abstract}
Embodied vision--language models must answer questions from partial, evolving observations. Their success depends on both the reasoning capability selected and the evidence available to it. Expert routers adapt internal computation, while tool-augmented systems acquire external evidence; choosing these independently can leave a suitable expert without the evidence it needs.
We introduce \method, a bounded framework that jointly selects a capability expert, optional knowledge, and perceptual tools. Outcome-derived signatures represent the overlapping strengths of General, Spatial Reasoning, and Spatiotemporal experts. A joint utility scores complete expert--evidence routes under a call budget. After acquisition, a packet verifier admits valid evidence with positive predicted expert-conditioned gain. The selected expert then answers once, reverting to the unchanged observation if all packets are rejected.
Across 6 embodied QA benchmarks, \method achieves a mean score of 61.4, the highest among the compared systems, and averages 74.7 on 10 general vision--language benchmarks. With all external evidence disabled, question-conditioned expert selection alone improves over the best fixed expert by 1.75 points. Ablations show that learned routing is the dominant component and that knowledge and perception are complementary.

\end{abstract}

\section{Introduction}
Embodied vision--language models (EVLMs) answer questions about evolving scenes from partial observations. Such grounded understanding supports language-guided robotic planning and interaction \citep{liang2025largemodelempoweredembodied}. Queries require
semantic knowledge, precise geometry, or temporal state tracking; decisive cues may occupy only a few pixels or span several frames \citep{majumdar2024openeqa,sermanet2024robovqa,yang2025thinking}. Success depends on both the selected capability and the evidence it receives. A relative-position question may benefit from localization, whereas an event-order question requires temporal context; neither source is useful merely because it is available. The route must therefore account for what the question demands, what each expert can infer from the original observation, and what an external provider can add.
We study \emph{expert--evidence alignment under partial observability}: matching a reusable capability with query-specific evidence, while retaining a generalist path and avoiding an unbounded agent loop.

Existing methods address different parts of this problem. Embodied generalists such as LEO and RynnBrain integrate perception, language, and action capabilities, whereas SpatialVLM, SpatialRGPT, and RoboSpatial use geometric supervision to strengthen spatial reasoning
\citep{huang2024leo,dang2026rynnbrain,chen2024spatialvlm,cheng2024spatialrgpt,song2025robospatial}.
Mixture-of-experts methods provide conditional parameter paths for efficient
specialization \citep{wu2024mole,shen2024mome}; ViperGPT and LLaVA-Plus instead compose external vision modules to recover information that a backbone cannot reliably infer \citep{suris2023vipergpt,liu2024llavaplus}. These expert routers focus on selecting a parameter path, whereas the cited tool-augmented systems acquire evidence for a fixed or implicit reasoner. Jointly choosing the capability path and its external evidence makes their dependence an explicit part of the routing decision.

These coupled choices pose two challenges. \textcolor{EKTAccent}{\textbf{(i)}}\,\textbf{Evidence value is expert-dependent.} Expert strengths overlap, and the same knowledge or tool output can provide different gains to different experts. Selecting the expert and evidence independently can therefore overlook useful combinations. \textcolor{EKTAccent}{\textbf{(ii)}}\,\textbf{Predicted evidence value need not match the usefulness of the returned packets.} A promising route may still yield incomplete or ambiguous evidence; deciding what to retain requires inspecting packet coverage, confidence, ambiguity, and provenance after execution. The challenge is to account for this newly available information without restarting expert selection or expanding the acquisition budget. Our design separates these uncertainties in time: select a complete expert--evidence route before acquisition, then revise only the evidence decision after inspecting the packets. Conditional evidence use thus requires neither repeated expert selection nor repeated answer generation.

\begin{figure}[t]
  \centering
  \includegraphics[width=\linewidth]{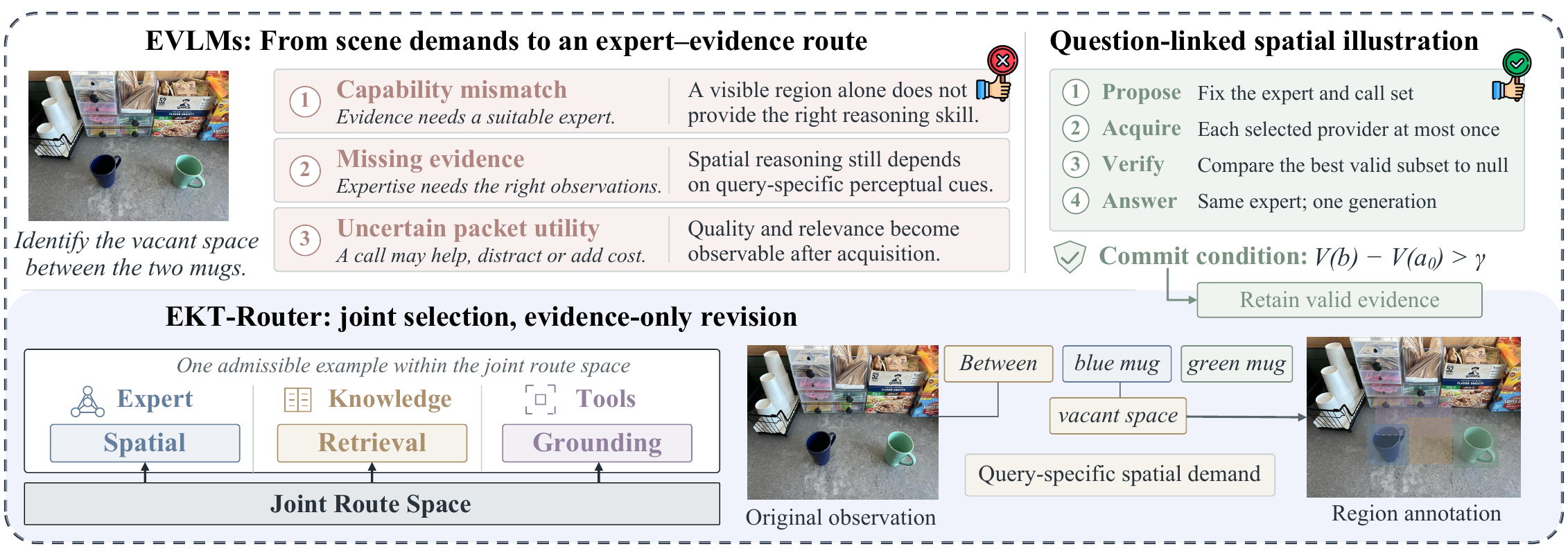}
\caption{\method{} couples capability selection with evidence acquisition. Top: three conceptual failure modes of independent decisions. Bottom: the Route Space includes one expert, optional knowledge, and a budgeted tool subset, followed by acquisition, packet verification, and one answer.} %
  \label{fig:motivation}
\end{figure}

\method{} implements this decision in two stages (Figure~\ref{fig:motivation}). Offline, independent capability post-training produces three Qwen3-VL-8B LoRA experts: General through continual pretraining (CPT), Spatial Reasoning through spatial supervision, and Spatiotemporal through video supervision. Diagnostic outcomes encode their overlapping strengths as continuous capability signatures, while paired executions supervise a complete-tuple route utility. Online, a frozen Qwen3-VL-2B provides a query-aware control view; the router selects one expert, optional Knowledge, and a budget-feasible Tool subset. Knowledge supplies task priors, while Tools return grounding, depth, or text as typed packets. A bounded verifier selects a valid evidence subset by its predicted
score margin over the null reference. The fixed expert answers once from the observation and retained evidence; rejecting every packet recovers the evidence-free path.
The two stages assign different roles to prediction and observation: the router estimates value before paying for evidence, while the verifier reassesses the returned packets after that cost is incurred. The null action belongs to both stages, so evidence use is optional throughout.
Our contributions are:

\noindent 1) We formulate expert--evidence alignment under partial observability as one decision coupling a capability path with optional knowledge and a budget-feasible tool set.

\noindent 2) We introduce \method{}, which derives continuous expert signatures and learns complete-route utility from answer outcomes. Typed packet verification refines evidence use after acquisition and preserves an exact evidence-free fallback.

\noindent 3) Across 16 benchmarks, \method{} achieves the best embodied QA average among the compared systems while matching strong general VLMs. Matched controls show that expert selection pays off even without external evidence, and ablations identify learned routing as the dominant component and knowledge and perception as complementary.

\section{Related Work}

\paragraph{Embodied question answering.}
Early work studied situated question answering and instruction-following navigation \citep{das2018eqa,anderson2018vln}. Foundation-model benchmarks now evaluate open-vocabulary episodic memory, spatial representation, and interactive or long-horizon reasoning \citep{majumdar2024openeqa,yang2025thinking,ni2025embodiedarena}. Spatially specialized VLMs improve geometric and grounded reasoning \citep{chen2024spatialvlm,cheng2024spatialrgpt}. The diversity of image, multi-frame, and video questions motivates measuring where specialists are complementary. Existing EQA models nevertheless remain limited in \textit{turning partial visual observations into reliable evidence for spatial
and cross-frame reasoning} \citep{majumdar2024openeqa,yang2025thinking}.

\paragraph{Routed parameter-efficient specialization.}
LoRA separates adaptation from frozen backbone weights \citep{hu2022lora};
routed extensions distribute examples or tokens across multiple low-rank paths.
Representative designs include hierarchical composition, anti-forgetting
plugin mixtures, token-level routing for conflicting data, and sparse
multimodal experts
\citep{wu2024mole,dou2024loramoe,chen2024llavamole,shen2024mome}.
Continual and embodied variants further allocate adapters by instruction, layer,
or task \citep{ge2025dynamic,jia2025hierarchical,luo2026coral}. RouteLLM learns
model selection from preferences \citep{ong2025routellm}.
Route-to-Reason jointly selects a model and a reasoning strategy under a budget \citep{pan2025route}, and MASER routes spatial questions among modality adapters on a shared embodied backbone \citep{raj2026maser}.
These methods select models, parameter paths, or internal strategies.
\textit{\method{} instead scores a complete route: a capability path together with the non-parametric evidence that the path will consume.}

\paragraph{Tool-augmented multimodal reasoning.}
Modular systems recover missing information through programs, retrieval, or vision APIs \citep{suris2023vipergpt,lu2023chameleon,gao2023assistgpt}. LLaVA-Plus likewise composes external perception modules \citep{liu2024llavaplus}. Recent work develops spatial toolkits \citep{ruan2026world2mind}, call-level credit assignment \citep{feng2026taco}, self-regulated visual tools \citep{chen2026bee}, and latent control for agent decisions \citep{ghasemabadi2026latent}.
\textit{Beyond selecting tools, \method{} distinguishes between acquiring evidence and accepting it, with its utility evaluated for the selected expert.}

\section{Method}
\label{sec:method}

\begin{figure}[t]
  \centering
  \includegraphics[width=\linewidth]{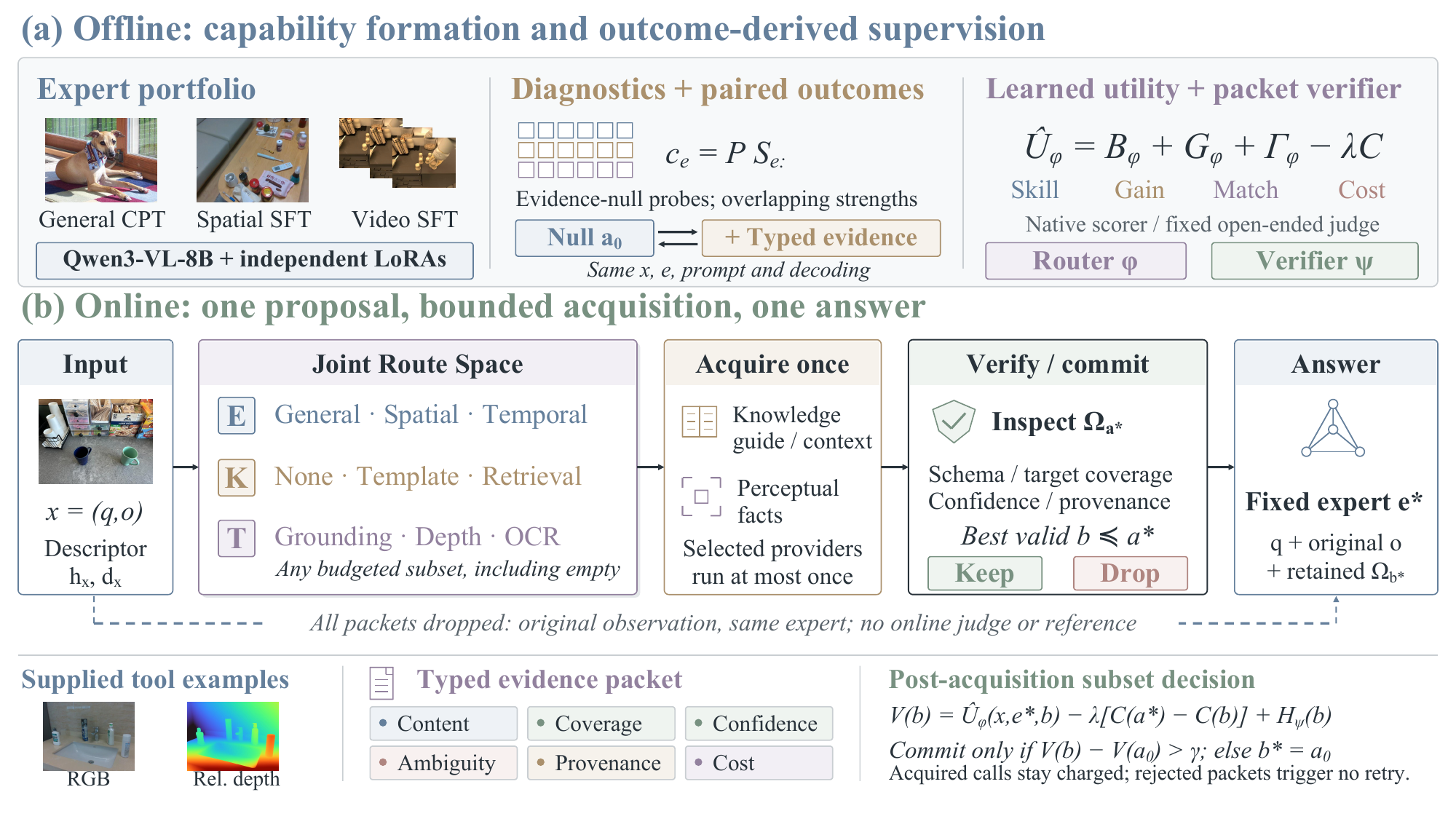}
\caption{\textbf{Overview of \method{}.}
\textit{(a) Offline:} independent adaptations form the expert portfolio;
evidence-free diagnostics yield capability signatures, and paired answer
outcomes supervise the router and verifier.
\textit{(b) Online:} $\phi$ jointly selects an expert and budgeted evidence actions.
Each selected provider runs at most once. $\psi$ scores valid packet subsets
against the null reference, retaining evidence only above margin $\gamma$.
The fixed expert answers once from the original observation and retained
evidence; the dashed path preserves the evidence-free fallback.}

\label{fig:framework}
\end{figure}

\method{} makes a two-stage decision under partial information. The Utility
Router first selects an expert and typed Knowledge and Tool actions. Once the packets return, the Packet Verifier revises the evidence decision on the same utility scale, keeping the expert fixed. Verification uses packet attributes, not candidate answers; only the final expert generates an answer. Figure~\ref{fig:framework} connects this online procedure to offline supervision. %

\subsection{Joint Expert--Evidence Formulation}

Given $x=(q,o)$, \method{} answers question $q$ from a partial visual
observation $o$ (an image, multiple frames, or a video) by jointly selecting
a reasoning expert and its external evidence. We denote the answer by
$\widehat y$ and the training reference by $y$. %

We form each expert by applying an independently trained LoRA update to the
shared Qwen3-VL-8B backbone with parameters $\theta_0$:
\begin{equation}
\Theta_e=\theta_0+\Delta_e,
\qquad e\in\mathcal E=\{\mathrm{gen},\mathrm{spa},\mathrm{st}\},
\label{eq:experts}
\end{equation}
where $\Delta_e$ is zero on unadapted weights. General uses continued
pretraining; Spatial Reasoning and Spatiotemporal use spatial
and video supervision. Their measured strengths overlap rather than form a
hierarchy. Routing activates one adapter, with all expert parameters frozen.

Knowledge providers $\mathcal K$ supply reasoning templates or retrieved text; Tools $\mathcal T$ extract scene evidence through grounding, depth estimation, or OCR. With $\mathcal K_0=\mathcal K\cup\{\emptyk\}$ allowing no knowledge, the joint route space is
\begin{equation}
\begin{aligned}
z&=(e,a)\in\mathcal E\times\mathcal A,\qquad a=(k,S),\\
\mathcal A&=\{(k,S):k\in\mathcal K_0,\ S\subseteq\mathcal T,
\ C_T(S)\leq B_T\}.
\end{aligned}
\label{eq:action-space}
\end{equation}
Here $S$ is a possibly empty tool subset with cost budget $B_T$, and
$C(a)=C_K(k)+C_T(S)$ with zero null costs. The null action
$a_0=(\emptyk,\varnothing)$ preserves the original-input path for every expert.

Selected non-null providers, indexed by $\mathcal I(a)$, return evidence packets:
\begin{equation}
\begin{aligned}
\Omega_a&=\{\omega_j\}_{j\in\mathcal I(a)},\qquad \Omega_{a_0}=\varnothing,\\
\omega_j&=(v_j,q_j^{\mathrm{cov}},q_j^{\mathrm{conf}},q_j^{\mathrm{amb}},p_j,C_j),
\end{aligned}
\label{eq:packets}
\end{equation}
where the fields record content, coverage, confidence, ambiguity, provenance, and cost. This separation keeps provider-specific content distinct from the attributes used to assess it. For example, grounding coverage identifies which target a region refers to, while depth provenance identifies the observation from which relative distances were estimated. A well-formed packet can still be irrelevant to the question; structural validity and predicted gain therefore answer different questions. The verifier inspects acquired packets before deterministic verbalization. Thus routing chooses what to acquire for an expert; verification decides which returned evidence that fixed expert receives.

\subsection{Outcome-Derived Expert Signatures}

A categorical expert name cannot express overlapping strengths. We instead
evaluate each expert without external evidence on a diagnostic collection
partitioned into $D$ capability probes $\{\mathcal G_d\}_{d=1}^{D}$. After
normalizing scores within each probe, we form
\begin{equation}
S_{ed}=\frac{1}{|\mathcal G_d|}\sum_{i\in\mathcal G_d}
\widetilde R_d\!\left(y_i,\widehat y_i(e,a_0)\right),
\label{eq:signature-score}
\end{equation}
and define the capability signature as $c_e=P S_{e:}\in\mathbb R^r$, where $P$ is a learned low-rank projection. Probe-wise normalization makes accuracy, exact match, and answer-judge scores comparable despite their different numerical ranges.

The frozen query encoder maps $x$ to $h_x$; a learned router mapping predicts demand $d_x$. We define the evidence-free expert score as
\begin{equation}
B_\phi(x,e)=d_x^\top W_Bc_e+b_e.
\label{eq:base-score}
\end{equation}
Expert signatures thus derive from diagnostic outcomes, not hand-written levels. Extending the portfolio requires new diagnostics and fitting the affected routing components. General, Spatial Reasoning, and Spatiotemporal remain descriptive names for the current paths.

\subsection{Interaction-Factorized Route Utility}

We score a route through three distinct terms:
\begin{equation}
\widehat U_\phi(x,e,a)=B_\phi(x,e)+G_\phi(x,a)
+\Gamma_\phi(x,e,a)-\lambda C(a).
\label{eq:utility}
\end{equation}
Here $B$ measures evidence-free competence, $G$ is an action value shared
across experts, and $\Gamma$ captures input-dependent compatibility between
expert and evidence. We anchor
$G_\phi(x,a_0)=\Gamma_\phi(x,e,a_0)=0$ and impose the centering
$\sum_{e\in\mathcal E}\Gamma_\phi(x,e,a)=0$, so $G$ is the expert-mean action
value and $\Gamma$ the expert-specific deviation from it.

Let $r_a$ be a learned typed-action embedding with $r_{a_0}=0$. We set
$G_\phi(x,a)=d_x^\top W_Gr_a$, then form a raw interaction and subtract its
expert mean:
\begin{equation}
\begin{aligned}
\overline\Gamma_\phi(x,e,a)
&=d_x^\top W_\Gamma\big[(W_Ec_e)\odot(W_Ar_a)\big],\\
\Gamma_\phi(x,e,a)
&=\overline\Gamma_\phi(x,e,a)
-\frac{1}{|\mathcal E|}\sum_{e'\in\mathcal E}
\overline\Gamma_\phi(x,e',a).
\end{aligned}
\label{eq:interaction}
\end{equation}
Here $\odot$ denotes element-wise multiplication. Without $\Gamma$, evidence
value is independent of the expert. With $\Gamma$, the same grounding action
can have different predicted value for Spatial Reasoning and Spatiotemporal.
We define acquisition gain relative to the selected expert's evidence-free route:
\begin{equation}
\widehat\Delta_\phi(x,e,a)=
\widehat U_\phi(x,e,a)-\widehat U_\phi(x,e,a_0).
\label{eq:predicted-gain}
\end{equation}
Before acquisition, this expert-dependent margin estimates evidence value
without observing the quality of the returned packets. The centering constraint also fixes an interpretation for the decomposition: a channel that helps all experts similarly is represented mainly by $G$, while departures from that shared effect enter $\Gamma$. This is a modeling constraint on the utility scores, rather than a claim that the data uniquely identify separate causal effects for the two terms.

\subsection{Route Supervision from Answers}

EQA datasets provide answers but no route labels. For training example $i$, we execute selected route pairs under identical prompts and decoding, using a benchmark-native scorer whenever one exists and a fixed answer judge only for open-ended evaluation. The normalized realized utility is
\begin{equation}
\begin{aligned}
u_i(e,a)&=\widetilde R_i\!\left(y_i,\widehat y_i(e,a)\right)-\lambda C(a),\\
\delta_i(e,a)&=u_i(e,a)-u_i(e,a_0).
\end{aligned}
\label{eq:realized-utility}
\end{equation}
For a fixed expert, $\delta_i$ measures the gain from evidence; null executions compare expert capabilities without external evidence.

Using $u_i(z)=u_i(e,a)$ and
$\widehat U_\phi(x,z)=\widehat U_\phi(x,e,a)$ for $z=(e,a)$, let
$\mathcal P_i$ contain ordered route pairs $(z^+,z^-)$ for which
$u_i(z^+)>u_i(z^-)+\epsilon$. We optimize the Bradley--Terry preference loss
\citep{bradley1952rank}
\begin{equation}
\mathcal L_{\mathrm{pref}}=-\sum_i
\sum_{(z^+,z^-)\in\mathcal P_i}
\log\sigma\!\left(\frac{\Delta\widehat U_i}{\tau}\right),
\quad
\Delta\widehat U_i=
\widehat U_\phi(x_i,z^+)-\widehat U_\phi(x_i,z^-),
\label{eq:preference}
\end{equation}
with a lower-weight Huber term $\mathcal L_{\mathrm{cal}}$ between predicted and realized gains. The full objective $\mathcal L=\mathcal L_{\mathrm{pref}}+\alpha\mathcal L_{\mathrm{cal}}$ combines route ranking with acquisition-gain supervision. A held-out routing split selects $\epsilon$, $\lambda$, $\alpha$, and the decision margin. Pairwise comparisons focus supervision on routes with distinguishable outcomes; the gain term supplies complementary information about the scale of improvement over the same expert without evidence. Keeping the experts frozen makes these route outcomes stationary with respect to the answer parameters during controller fitting.

Packet outcomes supervise a residual correction for non-null evidence.
For a proposed action $a$ and valid retained subset $b\preceq a$ with $b\neq a_0$, define
\begin{equation}
\rho_i(b\mid a)=
\widetilde R_i\!\left(y_i,\widehat y_i(e,b)\right)-\lambda C(a)
-\widehat U_\phi(x_i,e,b)+\lambda[C(a)-C(b)].
\label{eq:packet-target}
\end{equation}
Thus, $\rho_i(b\mid a)=u_i(e,b)-\widehat U_\phi(x_i,e,b)$ measures the discrepancy between realized and predicted route utility. We train $H_\psi(h_x,c_e,r_b,\Omega_b)$ to predict this residual using a Huber loss over valid non-null subsets, while fixing $H_\psi(h_x,c_e,r_{a_0},\varnothing)=0$ for the evidence-free reference. The resulting retention margin tracks realized evidence gain when $B_\phi$ is calibrated to evidence-null outcomes (Appendix~\ref{app:method-details}).

\subsection{Bounded and Verified Inference}

At inference, the router first proposes
\begin{equation}
(e^\star,a^\star)=\arg\max_{e\in\mathcal E,\,a\in\mathcal A}
\widehat U_\phi(x,e,a).
\label{eq:proposal}
\end{equation}
One router pass scores all discrete tuples without generating candidate answers. Only the non-null components of $a^\star$ execute, each at most once, to produce $\Omega_{a^\star}$.

A validity predicate checks schema consistency, target coverage, finite
confidence, and provenance; the null subset is always valid. Write
$b=(k',S')\preceq a^\star=(k^\star,S^\star)$ when
$k'\in\{\emptyk,k^\star\}$ and $S'\subseteq S^\star$. Valid subsets are then
rescored after acquisition:
\begin{equation}
\widehat V_{\phi,\psi}(b\mid x,e^\star,a^\star)=
\widehat U_\phi(x,e^\star,b)-\lambda[C(a^\star)-C(b)]
+H_\psi(h_x,c_{e^\star},r_b,\Omega_b).
\label{eq:verified-score}
\end{equation}
The cost adjustment fixes incurred cost at $C(a^\star)$, which cancels across
valid subsets. Acquisition cost matters when choosing a route, but is already
sunk when deciding which returned packets to retain. The valid maximizer is
committed only if its predicted score margin over the null subset exceeds $\gamma$;
otherwise $b^\star=a_0$. Knowledge and Tool packets can be rejected independently.

Finally, the fixed expert answers from the original observation and retained
packets:
\begin{equation}
\widehat y\sim p_{\Theta_{e^\star}}
\!\left(\cdot\mid q,\,o,\,\operatorname{verbalize}(\Omega_{b^\star})\right).
\label{eq:generation}
\end{equation}
Empty packets add no text, so $b^\star=a_0$ recovers the same expert's
evidence-free path. Verification generates no candidate answer, each provider runs at most once, and the answer is generated once under
$\mathbf 1[k^\star\neq\emptyk]+|S^\star|\leq B_{\mathrm{ext}}$.

\paragraph{Structural guarantee.}
Define the induced provider-call bound as $B_{\mathrm{ext}}=\max_{(k,S)\in\mathcal A}
\{\mathbf 1[k\neq\emptyk]+|S|\}$.
Inference uses one route proposal, at most $B_{\mathrm{ext}}$ provider invocations, one subset-verification pass, and one answer generation. If no valid subset exceeds the null score by $\gamma$, then $b^\star=a_0$ and
Equation~\ref{eq:generation} recovers the selected expert's original-input path. This follows from null validity, finite subset search, and a fixed expert. General remains available through $(\mathrm{gen},a_0)$.

\section{Experiments}
\label{sec:experiments}

We design experiments to answer four questions. \textbf{Q1:} Does expert--evidence alignment improve embodied QA over strong general, embodied, and proprietary models? \textbf{Q2:} Are the experts complementary, and does routing among them help without any external evidence? \textbf{Q3:} Which decisions (routing, knowledge, perception, supervision) account for the gain? \textbf{Q4:} Does specialization come at the cost of general vision--language competence?

\subsection{Evaluation Protocol}
The primary suite comprises six embodied QA benchmarks spanning single-image, multi-frame, and video inputs: UniEQA \citep{tjurl2026unieqa}, OpenEQA \citep{majumdar2024openeqa}, VSI-Bench \citep{yang2025thinking}, ERQA \citep{gemini2025robotics}, RoboVQA \citep{sermanet2024robovqa}, and Where2Place \citep{yuan2024robopoint}. We retain each native higher-is-better metric and report the equal-weight mean over the six scores; a separate ten-benchmark suite assesses general reasoning, document understanding, and hallucination.

We compare with Qwen3-VL, InternVL3.5, and the embodied models MiMo-Embodied and Pelican-VL \citep{bai2025qwen3vl,wang2025internvl35,hao2025mimo,zhang2025pelican}; proprietary models use their respective endpoints.
The three experts share frozen Qwen3-VL-8B and identical decoding, with one independently trained LoRA checkpoint active per query; General is CPT-trained, not the unadapted backbone. The descriptor and experts stay frozen during router training, and only the media-disjoint routing split supervises the router.

Knowledge templates cover reference-frame normalization, spatial decomposition, and temporal ordering. Hybrid retrieval returns at most five passages from a capability-indexed corpus. Tools include Grounding DINO \citep{liu2024groundingdino}, Depth Anything V2 \citep{yang2024depthanything}, and OCR; any budget-feasible subset may be selected. Packets retain coordinates, frame indices, text, confidence, and provenance before joint verbalization for one fixed-expert answer.

\paragraph{Training and comparison protocols.}
The spatial and temporal adapters use scale 256, dropout 0.05, and two epochs of AdamW with learning rate $10^{-4}$, weight decay 0.1, cosine scheduling, and 5\% warm-up. A media-disjoint 1,020/180 split fixes early stopping, costs, margins, and commitment. Matched runs share prompts, zero-temperature
decoding, and generation budgets; native scorers are used when available, and the frozen answer judge sees no route identity. Decision ablations reuse checkpoints, instances, decoding, and scorers; training controls change the named training stage.
The six-benchmark comparison combines reproduced, official API, and reported results. Matched controls isolate contributions under a shared evaluation protocol, while cross-system comparisons describe overall performance. %

\begin{figure}[t]
  \centering
  \includegraphics[width=\linewidth]{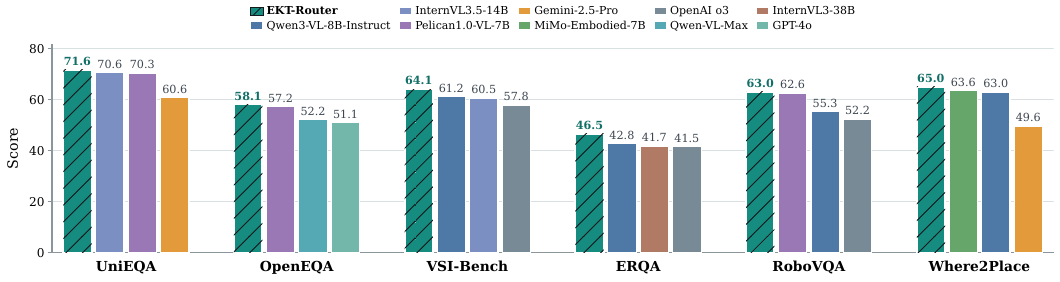}
  \caption{\textbf{Embodied QA across six benchmarks (Q1).} Each panel uses its native higher-is-better metric. Results combine reported and reproduced scores under the respective evaluation protocols.} %
  \label{fig:embodied-results}
\end{figure}

\begin{table}[t]
\caption{\textbf{General vision--language evaluation (Q4).} All metrics are higher-is-better. \textbf{Bold} identifies EKT-Router; \underline{underlining} marks column maxima, including ties. Both may apply. MMB: MMBench v1.1; MMS: MMStar; MathV: MathVista-mini; TVQA: TextVQA; CQA: ChartQA; DVQA: DocVQA; HallB: HallusionBench.} %
\label{tab:general-vlm}
\centering
\small
\setlength{\tabcolsep}{1.9pt}
\renewcommand{\arraystretch}{1.08}
\begin{tabular}{@{}lrrrrrrrrrrr@{}}
\toprule
& \multicolumn{5}{c}{Visual reasoning}
& \multicolumn{3}{c}{OCR / documents}
& \multicolumn{2}{c}{Hallucination} & \\
\cmidrule(lr){2-6}\cmidrule(lr){7-9}\cmidrule(lr){10-11}
Model & MMB & MMS & MMMU & AI2D & MathV & TVQA & CQA & DVQA & HallB & POPE & Avg. \\
\midrule
\rowcolor{black!9}
\multicolumn{12}{l}{\emph{Open general VLMs}} \\
LLaVA-OV-7B & 80.8 & 61.7 & 48.8 & 82.0 & 63.2 & 77.6 & 80.0 & 87.5 & 47.3 & 87.5 & 71.6 \\
MiniCPM-V2.6 & 78.0 & 57.5 & 49.8 & 82.1 & 60.6 & 80.1 & 79.3 & 90.5 & 48.1 & 87.0 & 71.3 \\
Cambrian-8B & 75.9 & 57.6 & 42.7 & 79.7 & 49.0 & 71.7 & 73.3 & 77.8 & 41.6 & 86.5 & 65.6 \\
InternVL2-8B & 79.4 & 61.5 & 51.2 & 83.8 & 58.3 & 77.4 & 83.3 & 91.6 & 45.0 & 86.4 & 71.8 \\
InternVL2.5-8B & \underline{82.2} & 62.8 & 56.0 & \underline{84.5} & \underline{64.4} & 78.2 & \underline{84.7} & 92.9 & 49.3 & \underline{88.5} & 74.4 \\
Qwen2-VL-7B & 80.5 & 60.7 & 54.1 & 83.0 & 58.2 & \underline{84.3} & 83.0 & 93.9 & 42.4 & 87.3 & 72.7 \\
Qwen3-VL-8B & 74.9 & 42.9 & 47.0 & 46.7 & 32.8 & 83.1 & 83.5 & \underline{95.8} & \underline{51.3} & 88.3 & 64.6 \\
\addlinespace[2pt]
\rowcolor{black!9}
\multicolumn{12}{l}{\emph{Embodied and spatial specialists}} \\
SpatialBot-3B & 58.2 & 39.1 & 31.4 & 56.7 & 35.0 & 50.2 & 48.1 & 56.3 & 36.2 & 81.5 & 49.3 \\
SpatialRGPT-7B & 72.3 & 52.8 & 43.5 & 74.6 & 49.1 & 70.5 & 68.4 & 78.0 & 41.0 & 85.7 & 63.6 \\
ManipVQA-7B & 69.5 & 50.2 & 40.3 & 71.5 & 46.8 & 67.0 & 65.3 & 75.2 & 38.8 & 83.7 & 60.8 \\
LEO-7B & 70.8 & 51.4 & 41.7 & 72.0 & 47.3 & 68.8 & 66.7 & 76.5 & 39.5 & 84.9 & 62.0 \\
\rowcolor{EKTRow}
\textbf{EKT-Router (8B)} & \textbf{81.9} & \underline{\textbf{63.2}} & \underline{\textbf{57.8}} & \textbf{84.3} & \textbf{61.5} & \textbf{83.9} & \textbf{84.2} & \textbf{93.8} & \textbf{47.9} & \textbf{88.3} & \underline{\textbf{74.7}} \\
\bottomrule
\end{tabular}

\end{table}

\subsection{Q1: Embodied QA Results}

\textbf{Finding 1: an 8B router surpasses larger and proprietary systems.} \method{} attains the highest six-benchmark average, 61.4, ahead of embodied specialists (MiMo-Embodied, Pelican-VL), general models up to 38B, and proprietary APIs including Gemini-2.5-Pro and o3, while activating a single 8B expert per query. It ranks first on all six benchmarks, leading the runner-up by +0.4 to +3.7 points, with the largest margins on ERQA (46.5 vs.\ 42.8) and VSI-Bench (64.1 vs.\ 61.2). Against its own General path under the matched protocol, \method{} gains +1.78 on average, driven by RoboVQA (+7.78), ERQA (+3.75), VSI-Bench (+2.94), and Where2Place (+2.00). These tasks require temporal state tracking, multiview grounding, and spatial affordance, precisely where a specialist supplied with grounding or depth packets can resolve cues that a generalist sees but cannot localize.

The gains are selective. On UniEQA ($-3.37$) and OpenEQA ($-2.43$), dominated by open-vocabulary semantics and episodic recall, the General path remains stronger; such queries benefit little from geometric evidence, which is why General is kept as an explicit route. Q2 and Q3 decompose the aggregate gain into expert selection and evidence use.

\subsection{Q2: Expert Complementarity}

\begin{figure}[t]
  \centering
  \includegraphics[width=\linewidth]{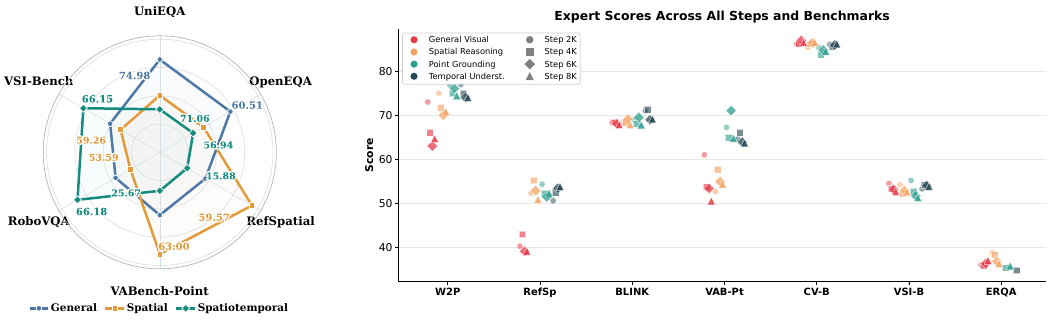}
  \caption{\textbf{Expert complementarity (Q2).} \textit{Left:} the three deployed experts reverse rank on six representative benchmarks from the matched control. \textit{Right:} candidate-checkpoint differences persist across training. Deployment retains General, Spatial Reasoning, and Spatiotemporal; the diagnostic Pointing/Grounding candidate becomes a selectable Tool action. W2P: Where2Place; RefSp: RefSpatial; VAB-Pt: VABench-Point; CV-B: CV-Bench; VSI-B: VSI-Bench.}

  \label{fig:complementarity}
\end{figure}

\textbf{Finding 2: no single expert dominates.} In Figure~\ref{fig:complementarity}, General leads UniEQA and OpenEQA, Spatial Reasoning leads RefSpatial and VABench-Point, and Spatiotemporal leads RoboVQA and VSI-Bench; these rank reversals persist from 2K to 8K steps rather than converging to one best checkpoint. Specialization thus creates stable, query-dependent complementarity, the precondition for routing to pay off. The diagnostic also shapes the portfolio: the Pointing/Grounding candidate overlaps with Spatial Reasoning, so localization is exposed as a Tool that any expert can consume. Unlike expert-only routers such as MoLE or MASER, evidence useful to several experts is shared as an action rather than duplicated as another expert.

\textbf{Finding 3: routing alone converts complementarity into accuracy.} To separate expert choice from evidence, we disable Knowledge and Tools for every candidate on nine benchmarks (the six primary tasks plus Cosmos, RefSpatial, and VABench-Point). Question-conditioned selection reaches 60.24 versus 58.49 for the best fixed expert, a +1.75 gain whose media-grouped 95\% CI $[1.25,2.26]$ excludes zero. With no external evidence involved, this gain is attributable to the router: signatures and utility recover which expert suits each query, which no globally best model can.

\subsection{Q3: Mechanism Attribution and Sensitivity}
\begin{figure}[t]
  \centering
  \includegraphics[width=\linewidth]{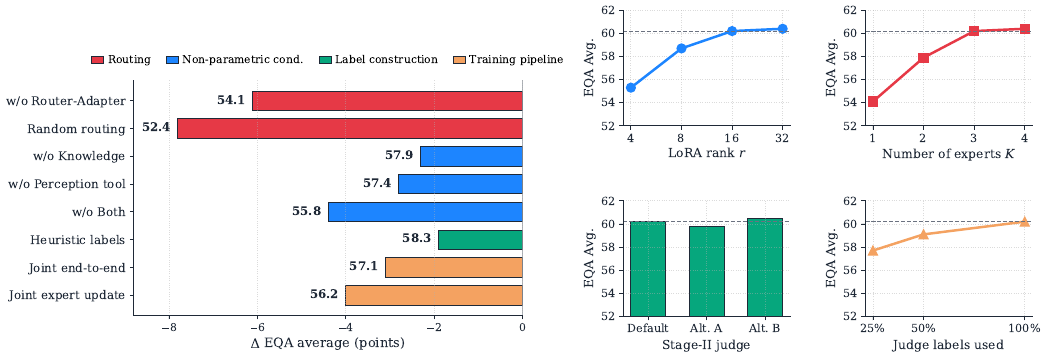}
  \caption{\textbf{Ablations and sensitivity under the separate frozen protocol (Q3).}
  \textit{Left:} changes from the 60.2 full-system score. \textit{Right:} LoRA rank, expert count, answer judge, and route-label coverage. Dashed lines mark the full system.} %
  \label{fig:ablation-sensitivity}
\end{figure}

\textbf{Finding 4: learned routing is the dominant component.} Figure~\ref{fig:ablation-sensitivity} (left) ablates each decision relative to the 60.2 full system. Removing the router causes the largest drop ($-6.1$, to 54.1), and random routing is worse still ($-7.8$, to 52.4), 1.7 points below using no router at all. Owning several adapters is thus harmful when they are mismatched to queries; the portfolio pays off only when a learned utility assigns each query to the right expert. These within-protocol controls support query-dependent assignment over mere portfolio availability. They do not separately identify every interaction term in the utility model, and their absolute scores should not be subtracted from the Q1 comparison.

\textbf{Finding 5: knowledge and perception are super-additive.} Removing Knowledge ($-2.3$), perception ($-2.8$), or both ($-4.4$) all hurt. From the evidence-free variant (55.8), restoring both channels recovers 4.4 points, 0.7 more than the sum of single-channel recoveries (1.6 + 2.1). The channels fix different failures: Knowledge supplies procedures such as reference-frame normalization, whereas perception supplies observation-bound measurements. A router that can request both for the same expert exploits this synergy, which expert-only routing cannot access.

\textbf{Finding 6: outcome supervision beats hand-written maps and joint training.} Heuristic capability labels score 58.3 ($-1.9$), so measured expert strengths are more informative than intuitive task-to-expert assignments. Joint end-to-end training (57.1, $-3.1$) and joint expert updates (56.2, $-4.0$) are both worse than freezing the experts: updating experts while learning the router shifts the targets the router ranks. Decoupling capability formation from route learning is thus a deliberate design choice, not merely a cost saving.

Figure~\ref{fig:ablation-sensitivity} (right) shows saturation rather than fragility. In a diagnostic sweep, accuracy rises from 55.3 at LoRA rank 4 to 60.2 at rank 16. Three experts improve over two by 2.3 points, while a fourth adds 0.2, matching the three capability regimes in Q2. Scores stay within 59.8--60.5 across answer judges, and halving the route labels costs only 1.1 points (59.1), which indicates limited sensitivity within these reported settings. The sweep does not establish robustness to arbitrary provider failures or distribution shifts, and the rank setting belongs to this diagnostic protocol rather than the main expert-adaptation recipe.

\paragraph{Selective evidence acquisition.}

Each channel helps where its semantics match the query: Knowledge supplies procedures, grounding exposes small entities, depth resolves relative distance, and OCR recovers answer-bearing text. The same channels can also hurt, through scene-incompatible retrieval, repeated-instance confusion, or over-weighted low-confidence measurements. This is why \method{} treats acquisition and acceptance as separate decisions: the verifier can discard such packets after they return and fall back to the evidence-free path, whereas tool-augmented pipelines that always consume tool outputs inherit these errors.

\subsection{Q4: General Vision--Language Performance}

\textbf{Finding 7: competitive general vision--language performance.} Embodied and spatial specialists typically trade general competence for geometric skill: in Table~\ref{tab:general-vlm}, the strongest of them, SpatialRGPT-7B (63.6), trails InternVL2.5-8B (74.4) by 10.8 points. \method{} remains competitive in this cross-system comparison. It averages 74.7, on par with the strongest open general VLM, and achieves the best MMStar (63.2) and MMMU (57.8) among all models, surpassing InternVL2.5-8B by +0.4 and +1.8 on these two reasoning-heavy benchmarks. It exceeds SpatialRGPT-7B by 11.1 points on average and on every benchmark, from +2.6 on POPE to +15.8 on ChartQA and DocVQA, and the embodied generalist LEO-7B by 12.7 points. The large document and chart margins are consistent with retaining a separate generalist path: specialization lives in independent adapters, and the CPT-trained General expert remains available for non-embodied queries. Because Table~\ref{tab:general-vlm} compares different systems, however, it is not a matched before-and-after test of how much general ability each adaptation preserves. A single deployed model thus serves both embodied and general workloads.

\subsection{Inference Cost and Auditability}
\textbf{Bounded, query-adaptive execution.} Each query runs a fixed pipeline: one descriptor pass, one route proposal, one subset verification, and one answer generation. Unlike iterative agents, whose tool calls and generations grow with the reasoning trace, \method{} invokes at most $B_{\mathrm{ext}}$ providers and generates exactly one answer, regardless of query difficulty. Cost also adapts to demand: queries that need no evidence take the null route and trigger no retrieval or perception call, while the others invoke at most one Knowledge provider and a budget-feasible Tool subset.

\textbf{Lightweight and auditable control.} The answer path activates one frozen 8B backbone with a single expert adapter, and a frozen 2B descriptor supplies the control view. The only trainable controllers, the Utility Router and Packet Verifier, total about 1.245M parameters, less than 0.02\% of the answer model, which limits the additional parameter count. Parameter count alone does not establish end-to-end latency: descriptor execution, media encoding, provider initialization, retrieval, and adapter switching also contribute to runtime. Because every decision is explicit, logs record the descriptor, route utilities, selected tuple, packet provenance and confidence, and the answer, supporting stage-wise inspection of demand estimation, acquisition, compatibility, and generation. These records expose where a decision was made; determining the cause of an incorrect answer still requires matched comparisons and inspection of the underlying evidence.

\section{Conclusion}
Embodied QA needs a reasoning capability and evidence suited to the same query; choosing them independently can pair an expert with unusable evidence. \method{} aligns the two in one bounded route: outcome-derived signatures describe what each expert can do, a factorized utility predicts which evidence helps which expert, and packet verification decides what to keep after acquisition, with an exact evidence-free fallback. With one 8B expert active per query, \method{} achieves the best embodied QA average among the compared systems while matching strong general VLMs. Matched controls show that query-conditioned expert selection pays off on its own, and ablations show that learned routing is the dominant component and that knowledge and perception are complementary. Together, these comparisons support treating \textbf{which expert should see which evidence} as an explicit decision. They also delimit the conclusion: the gains depend on the expert portfolio, evidence interfaces, and evaluation protocol, while the bounded execution and original-input fallback are structural properties of the method. Appendix~\ref{app:limitations} outlines future directions.

\subsection*{AI use statement}
Generative AI tools assisted with literature discovery, language polishing, and LaTeX formatting. They did not conduct the reported experiments or generate the
reported measurements. The authors take responsibility for the final text,
claims, code, and results, including the verification of primary sources,
experimental artifacts, equations, citations, and AI-assisted edits.

{\small\setlength{\bibsep}{3pt plus 0.5pt}
\bibliography{references}
\bibliographystyle{plainnat}}

\appendix
\section{Expert--Evidence Routing Details}
\label{app:method-details}

This section details the information available at each decision stage, the
relation between acquisition and retention scores, and the finite search that implements Section~\ref{sec:method}. The selected expert and original observation remain fixed throughout evidence verification.

\subsection{Information Available to the Two Decisions}

The portfolio contains General, Spatial Reasoning, and Spatiotemporal experts.
Each is an independently adapted path on the shared backbone, and its label
describes its principal supervision rather than an exclusive task assignment.
Evidence-free probe outcomes in Equation~\ref{eq:signature-score} characterize overlapping strengths. Every expert can be selected with the null action; General can also be paired with Knowledge or Tools.

Before acquisition, the router observes the question and original visual
input through the frozen descriptor, together with expert signatures and
candidate action embeddings. Provider outputs are unavailable at this stage:
$\widehat U_\phi(x,e,a)$ predicts how useful an action will be for an expert.
After acquisition, the verifier additionally observes the returned packet
content and attributes. It chooses a subset of these packets without changing the expert or acquiring further evidence. Reference answers and answer-judge outputs are used for offline supervision, not either online decision.

\subsection{Packet Semantics and Valid Subsets}

A packet separates evidence content from coverage, confidence, ambiguity,
provenance, and cost. Coverage associates content with its target regions or
frames; provenance associates it with a provider and source observation or
passage. Confidence and ambiguity describe the evidence returned by that
provider. These attributes are inspected before deterministic verbalization,
so rejecting a packet removes its added evidence without altering the original observation.

For the proposed action $a^\star=(k^\star,S^\star)$, each retained action
$b=(k',S')$ satisfies $k'\in\{\emptyk,k^\star\}$ and $S'\subseteq S^\star$.
Retention operates on the provider packets defined in Equation~\ref{eq:packets}. Let
\begin{equation}
\mathcal B(a^\star)=
\left\{b\preceq a^\star:
\operatorname{valid}(b,\Omega_b)=1\right\},
\qquad a_0\in\mathcal B(a^\star).
\label{eq:appendix-valid-subsets}
\end{equation}
The deterministic predicate checks schema consistency, target coverage,
finite confidence, and provenance. The learned verifier ranks subsets that
pass these checks; validity and predicted utility therefore serve distinct
roles.

\subsection{Null Cost, Residual Training, and Retention Scores}

The cost convention is $C_K(\emptyk)=C_T(\varnothing)=C(a_0)=0$.
Selecting $a_0$ before acquisition incurs no external-provider cost. Selecting a non-null $a^\star$ and later retaining $a_0$ still incurs $C(a^\star)$, because the providers have already executed. This distinction is implemented by the adjustment in Equation~\ref{eq:verified-score}.

For a training proposal $a$ and a valid non-null subset $b\preceq a$, the
packet target in Equation~\ref{eq:packet-target} simplifies to
\begin{equation}
\begin{aligned}
\rho_i(b\mid a)
&=\widetilde R_i\!\left(y_i,\widehat y_i(e,b)\right)
  -\lambda C(b)-\widehat U_\phi(x_i,e,b)\\
&=u_i(e,b)-\widehat U_\phi(x_i,e,b).
\end{aligned}
\label{eq:appendix-residual}
\end{equation}
The two terms involving $C(a)$ cancel in this target. Thus $H_\psi$ learns the remaining discrepancy between realized and predicted utility for retained non-null evidence, while the inference score separately accounts for the cost of the original proposal.

Writing $\ell_{\mathrm{Huber}}$ for the Huber loss, the verifier objective has
the form
\begin{equation}
\begin{aligned}
\mathcal L_H
&=\sum_{(i,e,a,b)\in\mathcal D_H}
\ell_{\mathrm{Huber}}\!\left(
H_\psi(h_{x_i},c_e,r_b,\Omega_b)-\rho_i(b\mid a)\right),\\
\mathcal D_H
&=\{(i,e,a,b): b\preceq a,\ b\neq a_0,\
\operatorname{valid}(b,\Omega_b)=1\}.
\end{aligned}
\label{eq:appendix-verifier-loss}
\end{equation}
Here $\mathcal D_H$ ranges over the collected training executions.
The null correction is fixed to
$H_\psi(h_x,c_e,r_{a_0},\varnothing)=0$ and is excluded from the residual
loss. The null reference receives no learned residual correction.

Substituting Equation~\ref{eq:utility} into Equation~\ref{eq:verified-score}
gives
\begin{equation}
\begin{aligned}
\widehat V_{\phi,\psi}(b\mid x,e^\star,a^\star)
={}&B_\phi(x,e^\star)+G_\phi(x,b)+\Gamma_\phi(x,e^\star,b)\\
&+H_\psi(h_x,c_{e^\star},r_b,\Omega_b)-\lambda C(a^\star).
\end{aligned}
\label{eq:appendix-sunk-cost}
\end{equation}
Every retained subset, including $a_0$, is charged the same incurred cost.
Using the zero anchors for $G$, $\Gamma$, and $H$ at $a_0$, its predicted
margin over null is
\begin{equation}
\begin{aligned}
&\widehat V_{\phi,\psi}(b\mid x,e^\star,a^\star)
-\widehat V_{\phi,\psi}(a_0\mid x,e^\star,a^\star)\\
&\qquad=G_\phi(x,b)+\Gamma_\phi(x,e^\star,b)
+H_\psi(h_x,c_{e^\star},r_b,\Omega_b).
\end{aligned}
\label{eq:appendix-retained-gain}
\end{equation}
The router uses cost when choosing acquisition; the verifier uses returned content to choose what to retain. Shared acquisition cost and the null anchors yield this identity for the predicted score margin.

\paragraph{Null calibration.}
Write $R_b$ and $R_0$ for the normalized scores of the same query and expert with retained subset $b$ and with no evidence. If $H_\psi$ fits Equation~\ref{eq:appendix-residual} exactly, the margin in Equation~\ref{eq:appendix-retained-gain} equals $R_b-\lambda C(b)-B_\phi(x,e^\star)$. It matches the realized gain $R_b-\lambda C(b)-R_0$ only when the evidence-free score is calibrated, $B_\phi(x,e^\star)\approx R_0$. The preference loss alone constrains utility differences, not this absolute level; the gain term $\mathcal L_{\mathrm{cal}}$ and the held-out margin $\gamma$ absorb shared offsets but not query-dependent miscalibration of $B_\phi$. An equivalent formulation that removes this dependence trains $H_\psi$ on the paired residual $\delta_i(e,b)-\widehat\Delta_\phi(x_i,e,b)$, for which an exact fit gives the realized gain directly. Reporting held-out calibration of $B_\phi$ against evidence-null outcomes, per expert, is therefore part of verifying the retention rule.

\subsection{Finite Search and Original-Input Fallback}

Let $\mathcal S_T=\{S\subseteq\mathcal T:C_T(S)\leq B_T\}$.
The router scores $|\mathcal E|(|\mathcal K|+1)|\mathcal S_T|$ discrete
tuples before acquisition. If the selected route contains
$m=\mathbf 1[k^\star\neq\emptyk]+|S^\star|$ non-null providers, it has at
most $2^m$ retained subsets before validity filtering. The deployed menu of
three Tool types and at most one Knowledge provider therefore permits at most
four provider invocations and sixteen retained subsets per proposal; the
configured budget can reduce these bounds. Scoring an additional subset does
not invoke a provider or generate a candidate answer.

Algorithm~\ref{alg:appendix-inference} makes the two decisions explicit.
The null subset guarantees a defined fallback even when all non-null packets
fail validation. If no valid subset clears the commitment margin, the retained
packet set is empty and Equation~\ref{eq:generation} becomes
$p_{\Theta_{e^\star}}(\cdot\mid q,o)$, the same selected expert conditioned on
the original input. This identity preserves the input path independently of
which providers were proposed.

\begin{algorithm}[t]
\caption{Bounded expert--evidence inference}

\label{alg:appendix-inference}
\begin{algorithmic}[1]
\Require Query $x=(q,o)$, frozen experts, feasible actions $\mathcal A$, margin $\gamma$
\State Encode $x$ with the frozen query descriptor.
\State $(e^\star,a^\star)\gets\arg\max_{e\in\mathcal E,\,a\in\mathcal A}\widehat U_\phi(x,e,a)$
\State Execute each non-null provider in $a^\star$ once to obtain $\Omega_{a^\star}$.
\State Enumerate $\mathcal B(a^\star)$, including the always-valid null subset $a_0$.
\State Compute $v_b\gets\widehat V_{\phi,\psi}(b\mid x,e^\star,a^\star)$ for every $b\in\mathcal B(a^\star)$.
\State $\widehat b\gets\arg\max_{b\in\mathcal B(a^\star)}v_b$
\If{$v_{\widehat b}-v_{a_0}>\gamma$}
  \State $b^\star\gets\widehat b$
\Else
  \State $b^\star\gets a_0$
\EndIf
\State Generate one answer with $\Theta_{e^\star}$ from $q$, unchanged $o$, and $\operatorname{verbalize}(\Omega_{b^\star})$.
\end{algorithmic}

\end{algorithm}

\section{Training and Evaluation Details}
\label{app:training-evaluation}

\subsection{Adaptation and Routing Stages}

Expert adaptation precedes router fitting. General uses continued pretraining;
Spatial Reasoning and Spatiotemporal use spatial and video supervision,
respectively. The completed experts and the query descriptor remain frozen
during router fitting. Evidence-free diagnostic outcomes define the capability
profiles, while paired route executions provide answer-derived utility and
packet-correction targets. Table~\ref{tab:appendix-configuration} collects the
configuration used in the main text.

\begin{table}[t]
\caption{\textbf{Model, adaptation, and routing configuration.} Spatial and temporal
adaptation settings are separate from the diagnostic LoRA-rank sweep in Q3.}

\label{tab:appendix-configuration}
\centering
\small
\renewcommand{\arraystretch}{1.13}
\rowcolors{2}{EKTRow}{white}
\begin{tabular}{@{}p{0.23\linewidth}p{0.72\linewidth}@{}}
\toprule
\rowcolor{EKTAccent!16}
\textbf{Component} & \textbf{Setting} \\
\midrule
Answer portfolio & Shared Qwen3-VL-8B; one active expert adapter per query \\
General adaptation & Continued pretraining \\
Query descriptor & Frozen Qwen3-VL-2B \\
Spatial/temporal LoRA & Rank 128, scale 256, dropout 0.05 \\
Adaptation optimizer & AdamW; two epochs; learning rate $10^{-4}$; weight decay 0.1 \\
Adaptation schedule & Cosine decay with 5\% warm-up \\
Routing split & 1,020 fitting cases and 180 held-out cases; media-disjoint \\
Matched decoding & Shared prompts and generation budgets; temperature zero \\
Knowledge retrieval & Five passages; evaluation answers and held-out scene descriptions excluded \\
Perceptual providers & Grounding DINO, Depth Anything V2, and OCR \\
\bottomrule
\end{tabular}

\end{table}

For a fixed expert and input, paired evidence-on and evidence-null executions
use the same prompts and decoding. Equation~\ref{eq:realized-utility} combines
their native normalized scores with action costs. Route pairs separated by
more than $\epsilon$ supervise preference ranking, and the lower-weight Huber
term fits realized acquisition gains. The packet objective in
Equation~\ref{eq:appendix-verifier-loss} additionally uses the returned evidence
attributes to correct non-null route predictions. The held-out routing cases
set operating costs and margins; benchmark test scores do not select these
operating points. This ordering separates capability formation, utility fitting, and operating-point selection. Media-group separation is particularly important when several questions share a scene: a question-level split alone could expose the router to closely related observations on both sides of the validation boundary.

\subsection{Comparison Units}

Q1 reports the equal-weight mean of the six primary embodied QA benchmark
scores, using each benchmark's native higher-is-better metric. Q2 adds Cosmos,
RefSpatial, and VABench-Point to those six tasks and disables both external
evidence channels for every compared expert. Q4 uses the separate ten-task
general vision--language suite in Table~\ref{tab:general-vlm}. These suite means
aggregate benchmark scores, rather than pooling all questions into one accuracy.

The six-task full-system mean in Q1 is 61.40. In the nine-task evidence-null
control, question-conditioned selection reaches 60.24 versus 58.49 for the
best fixed expert, yielding the reported 1.75-point improvement. Q3 reports
component changes relative to its own 60.2 full-system reference under the
separate controlled protocol. The Q3 effects are within-protocol comparisons;
subtracting its absolute scores from Q1 or Q2 would combine different
evaluation settings.

Decision ablations hold expert checkpoints, instances, decoding, and evaluators
fixed; training and rank controls change the named adaptation configuration.
The Q2
media-grouped bootstrap interval, $[1.25,2.26]$, describes uncertainty in the
selection gain under that matched evaluation. It does not measure variation
across independently retrained portfolios. Cross-system comparisons combine
reproduced, official API, and reported results under their respective settings.

\subsection{Mechanism-Level Interpretation of Q3}
\label{app:mechanism-interpretation}

The Q3 controls are interpreted as interventions on distinct decision
interfaces. The routing intervention removes query-dependent demand matching
while preserving the expert bank; the full learned route's advantage over the random and no-router conditions therefore tests whether the assignment rule accounts for the gain. The evidence interventions preserve the selected expert and vary only the available information: Knowledge supplies procedures or
semantic priors, whereas perception supplies measurements tied to the current
observation. The observed 4.4-point recovery with both channels exceeds the
3.7-point sum of the single-channel recoveries by 0.7 points, consistent with
these information sources covering different failure modes.

The supervision controls provide a second attribution check. A heuristic route
does not receive answer-derived credit, while joint expert updates alter the
answer functions at the same time as the route model. The frozen-expert,
outcome-derived setting thus keeps the answer path fixed while learning the
route utility, making its comparison with these controls more interpretable.
This is an attribution rationale rather than a claim that the controls isolate
all optimization effects.

The reported component removals do not independently identify the interaction
term $\Gamma$ or the Packet Verifier. A fully isolated audit would reuse the
same expert checkpoints, route candidates, provider outputs, and decoding while
comparing additive utility without $\Gamma$, utility with $\Gamma$ but without
packet correction, and a sequential expert-then-evidence selector. Such a
comparison would distinguish expert--evidence compatibility from generic tool
use and post-acquisition filtering; until then, the present evidence supports
the interface-level interpretation above and not a separate causal estimate
for either unablated component.

\section{Evidence Interfaces and Interpretation}
\label{app:qualitative}

\subsection{Channel Semantics}

Knowledge and perception contribute different information. Reference-frame,
spatial-decomposition, and temporal-ordering templates supply reasoning
procedures, while retrieved passages supply semantic context. Perceptual
providers instead extract information tied to the observed image or frames.
The selected expert determines how these complementary inputs are used to
answer the query. Table~\ref{tab:appendix-evidence} gives representative
interpretation issues that can arise in each channel.

\begin{table}[t]
\caption{\textbf{Evidence channels and representative interpretation issues.}
The channels supply different information and require different consistency checks.}

\label{tab:appendix-evidence}
\centering
\small
\renewcommand{\arraystretch}{1.13}
\rowcolors{2}{EKTRow}{white}
\begin{tabular}{@{}p{0.17\linewidth}p{0.36\linewidth}p{0.38\linewidth}@{}}
\toprule
\rowcolor{EKTAccent!16}
\textbf{Channel} & \textbf{Information supplied} & \textbf{Representative interpretation issue} \\
\midrule
Knowledge & Reasoning procedures or retrieved semantic context & A passage may assume a reference frame or scene condition inconsistent with the observation. \\
Grounding & Target localization in an image or frame & Repeated instances or a small target may create ambiguous associations. \\
Relative depth & Relative-distance structure within the observation & Relative depth alone does not specify a calibrated metric distance. \\
OCR & Visible text associated with a region & A transcription or region association may be incorrect. \\
\bottomrule
\end{tabular}

\end{table}

The RGB and relative-depth panels in Figures~\ref{fig:motivation}
and~\ref{fig:framework} illustrate how an observation can produce a distinct
perceptual representation. Its packet retains source and target associations
so the verifier can examine that representation before it is added to the
answer context. These conceptual panels illustrate the interface; they do not provide a measured success rate for a particular packet or a model-generated answer. Relative-depth evidence can support ordinal comparisons, but metric distance requires additional calibration and cannot be inferred from the representation alone. The original visual input remains available whether the packet
is retained or rejected.

\subsection{Interpreting Acquisition and Retention}

An evidence comparison is conditioned on the same expert, question,
observation, and decoding. Its evidence-on and null-evidence conditions differ
in the added packet content, whereas comparing different experts changes the
reasoning path as well. The distinction matters for attribution: evidence
utility is assessed by the former comparison, and expert complementarity by
the latter under matched evidence conditions.

If a retained subset fails the validity predicate, it is excluded from the
candidate set. If a valid packet is irrelevant or conflicts with other
evidence, its usefulness must instead be assessed through the predicted
expert-conditioned margin. A retained packet can still be misunderstood by
the answer expert, and an empty retained set can still leave an underspecified
question. These cases distinguish packet validity, evidence usefulness, and
answer reasoning, which are separate stages of the route.

\section{Limitations and Future Work}
\label{app:limitations}

\method{} treats \emph{which expert should see which evidence} as a first-class decision. Its bounded execution makes the decision interfaces explicit, but does not guarantee correct perception, calibrated utility, or correct final answers. The following limitations delimit the present system and motivate extensions.

\paragraph{Portfolio coverage and transfer.}
The current portfolio contains three experts and three tools. Outcome-derived signatures summarize performance on the chosen diagnostic distribution, so their usefulness depends on probe coverage and score normalization. A new adapter requires fresh diagnostics and fitting of the affected controller components; a capability name alone is insufficient. A larger portfolio also expands the feasible route set and the number of outcomes needed to compare alternatives. Evaluating transfer to new scenes, tasks, adapters, and providers is necessary before treating the learned signatures as portable capability estimates.

\paragraph{Single-round observation limits.}
\method{} decides once which evidence to acquire. Rejecting a poor packet preserves the original input, but cannot reveal an object that is occluded or recover an event absent from the observed frames. Embodied agents can sometimes move, look again, and query new viewpoints. Extending the route utility to sequential decisions, where the verifier's judgement of returned packets informs the next acquisition or a change of viewpoint, would connect expert--evidence alignment with active perception and information-gathering policies, while retaining the explicit budget and fallback that make the current design auditable.

\paragraph{Outcome validity beyond question answering.}
We use answer outcomes as the supervision signal because EQA provides them at scale. Extending the formulation to task success, plan feasibility, or navigation would require outcomes and budgets defined for those settings. An answer-based utility is not a validated proxy for action success or physical safety, and packet validity does not establish that an action is safe. Routing experts and evidence for action generation, rather than answers, is a natural path from embodied understanding to embodied control.

\paragraph{Calibration and continual updates.}
An online answer is not automatically a labeled training outcome: updating signatures or utilities requires a reliable reference, task feedback, or an independently validated scorer. Provider confidence scales can also change across models and domains, so packet features and the null-reference calibration must be checked together. Continual updates would need to control feedback bias and preserve a stable evaluation set. The current experiments do not establish the stability of such online adaptation, nor do the component ablations isolate a separate causal contribution from every verifier and interaction term.

\end{document}